# LLMs Generate Kitsch

**Xenia Klinge** [1, 2]   **Stefan A. Ortlieb** [3]   **Alexander Koller** [1]
[1] Saarland University   [2] DFKI   [3] University of Bamberg
{xklinge|koller}@lst.uni-saarland.de   stefan.ortlieb@uni-bamberg.de

## Abstract

Large Language Models (LLMs) are increasingly used to generate pictures, texts, music, videos, and other works that have traditionally required human creativity. LLM-generated artifacts are often rated better than human-generated works in controlled studies. At the same time, they can come across as generic and hollow. We propose to resolve this tension by arguing that LLMs systematically generate *kitsch*, and that this is a consequence of the way in which they are trained. We also show empirically that readers perceive LLM-generated stories as kitschier, if we control for their definition of "kitsch". We discuss implications for the design of future studies and for creative tasks such as research and coding.

## 1 Introduction

The use of generative models for producing artifacts that have traditionally required human creativity is growing explosively. To name just a few examples, 24% of corporate press releases and 10% of job postings in 2024 were written with LLM support (Liang et al., 2025); half of all photos available on Adobe Stock in April 2025 – over 300 million – and 28% of all new music uploaded to Deezer in September 2025 was AI-generated (Kneschke, 2025; Deezer, 2025); and so was 30% of all Python code uploaded to Github in December 2024 (Daniotti et al., 2025). A recent analysis revealed that at least 50 papers accepted to NeurIPS 2025 contained hallucinated references, suggesting LLM use (Shmatko et al., 2026).

Generative AI has matured to a point where these artifacts are becoming hard to distinguish from human-produced ones, and where the consumers of these artifacts actually like them. Not only is the widespread real-world use of generative AI evidence of this; it has also been shown in a number of controlled academic studies (Zhao et al., 2023; Ayers et al., 2023; Porter and Machery, 2024).

*She walks the earth with grace and pride,*
*A beauty that cannot be denied,*
*With eyes that shine like stars above,*
*And lips that speak of endless love.*

*But though she wears a smile so sweet,*
*A broken heart doth lie beneath,*
*For in her chest a pain doth beat,*
*A love unrequited, without relief.*

*And so she walks with heavy heart,*
*A figure haunting in the dark,*
*For love, the sweetest of all art,*
*Can also leave a painful mark.*

Figure 1: An AI-generated poem in the style of Lord Byron, from Porter and Machery (2024).

At the same time, there is a broad public debate about "AI slop" (Judkis, 2024) – mass-generated content that is superficially beautiful but shallow – drowning out authentic human-produced content. The LLM-generated poem in Figure 1 is technically and stylistically coherent, but the mental imagery it evokes appears lifeless and generic. Benjamin (1927) once famously criticized the Surrealists for passing off their dream journals and automatic writings as poetry, abdicating the artist's core responsibility of having something original to say. A century later, it seems that LLMs have made Benjamin's worst nightmare come true.

In this paper, we offer a way to reconcile the attractiveness of LLM-generated artifacts with this sense of hollowness: We argue that LLMs systematically generate *kitsch*. The word "kitsch" describes works that have the superficial characteristics of art, but are optimized for mass appeal rather than artistic intent (Dorfles, 1969; Kulka, 1996). We will review some characteristics that philosophers of art have established for kitsch, and will argue that these characteristics are an almost necessary consequence of the way in which we train LLMs and other models of Generative AI. We will focus primarily on LLMs and the generation of art, i.e. works that emphasize enjoyment over utility; but our arguments apply equally to other modalities, and we will discuss implications for other creative activities, such as programming and research.

We will complement this discussion with an empirical study in which we establish, for the first time, that human readers perceive LLM-generated short stories as kitschier than human-written ones – if we control for the way in which they define "kitsch". At the same time, our subjects generally enjoyed the LLM-written stories more. This confirms our point that people can "like" LLM-generated artifacts even when they are actually kitsch; this has implications for the design of future studies on AI-supported creation.

## 2 What is kitsch?

The word "kitsch" first appeared around 1870 in Munich, where it was used by painters and art dealers as a label for "cheap artistic stuff" (Călinescu, 1987, p. 234). Today, "kitsch" has entered many modern languages, and its use is no longer limited to the domain of visual art. As a derogatory term, it can be applied equally to Neuschwanstein Castle, garden gnomes, pulp fiction, daily soaps, or overly sentimental music.

Key to the concept of kitsch is that kitsch emphasizes *mass appeal* over artistic value. Philosophers of art typically turn up their noses at kitsch. They use *kitsch* either as a synonym for bad or fake art (Dorfles, 1969; Scruton, 2014) or to distinguish art in general from "tasteless mass produced trash" (Pazaurek, 1912, p. 349). This highbrow attitude toward popular taste is closely linked to a new concept of art that emerged around the time the word kitsch became internationally successful. Recent work in psychology has taken a more neutral stance on the concept of kitsch, pointing out that the antagonism of kitsch and avant-garde reflects two modes of aesthetic appreciation that answer to basic needs for intimacy and autonomy (Ortlieb and Carbon, 2019a).

In order to ground our argument that LLMs generate kitsch, we will review three criteria for kitsch from the literature. These were originally proposed by Kulka (1996) and are still relevant in contemporary research.

### 2.1 Appeal to stock emotions

In order to fulfill its purpose of mass appeal, kitsch evokes emotions that are shared among a large proportion of the population. In the novel *The Unbearable Lightness of Being*, Kundera (1984) describes this connection as follows: "Kitsch causes two tears to flow in quick succession. The first tear says: How nice to see children running on the grass! The second tear says: How nice to be moved, together with all mankind, by children running on the grass! It is the second tear that makes kitsch kitsch."

As a consequence, kitsch typically depicts subjects or themes that will reliably "trigger an *unreflective emotional response*" (Kulka, 1996). The exact emotion that is triggered depends on the viewer's or reader's personal experience (Ortlieb et al., 2020); but kitsch thrives most on broad-strokes emotions to which many people can connect, such as love, family, friendship, nostalgia, patriotism, or religion – the "lowest common denominators of experience" (Greenberg, 1939).

Children running on the grass are symbolic of the happiness of childhood; the woman in the poem in Figure 1 is sad and in love. An original take of complex and nuanced emotions will not connect as universally with people and therefore damage the work's mass appeal; as Kulka says, kitsch addresses "stock emotions".

The emotional force of kitsch is furthermore *derivative of the emotional force of the subject it depicts*. People do not buy Eiffel Tower fridge magnets because they find artistic value in them; they buy them to remind themselves of how they felt in Paris. Similarly, the poem in Figure 1 derives much of its emotional impact from the fact that it is written in the style of Lord Byron, and this evokes the perceiver's prior emotional responses to actual works by Lord Byron and other Romantic poets. Kulka argues that the purpose of a work of kitsch is to symbolize the subject it depicts; in his words, "the appeal of kitsch is totally parasitic on the associations related to its referent".

### 2.2 Competent but conventional surface form

But if the depiction of an emotionally charged subject is the main purpose of a work of kitsch, this means that effective kitsch must make it as easy as possible for the perceiver to identify this subject.

Kulka stipulates that kitsch depicts its subject in a way that is "instantly and effortlessly identifiable". This has two important consequences for the surface form of kitsch.

First, a work of kitsch must be executed with *technical competence*. If we attempted to produce kitsch by having a toddler draw a replica of the Mona Lisa, the subject of the work would not be effortlessly identifiable, and therefore unsuitable as kitsch. Kitschy text must be coherent and grammatically correct; kitschy images must look plausible; kitschy music must involve reasonable chord progressions. Early models of generative AI could not ensure these properties. They were instantly recognizable as technically deficient, and therefore nobody seriously took their outputs for either art or kitsch.

Second, "kitsch invariably uses the most *conventional*, standard, well-tried, and tested representational canons" (Kulka, 1996). Imagine you wanted to produce kitsch around 1900. You would certainly not invent cubism as a painting style for your work, because this would confuse and perhaps offend your audience, damaging the mass appeal. Instead, kitsch plays it safe; it executes the current conventions competently, which makes it inherently backward-oriented (Greenberg, 1939).

In other words, just like kitsch is parasitic in terms of *content* on the emotional force of the subject it depicts, it is also parasitic in *style* on other works of art. Other artists must come up with stylistic innovations; once they become the established mainstream, kitsch can then adopt them.

### 2.3 Lack of artistic value

A third criterion is that kitsch is "something with the external characteristics of art, but which is in fact a falsification of art" (Dorfles, 1969): it looks like art, but "deceive[s] the consumer into thinking he feels something deep and serious, when in fact he feels nothing at all" (Scruton, 2014).

It seems counterintuitive at first glance that kitsch would be "false" art, when works of kitsch can in fact be quite beautiful to look at – their purely *aesthetic value* can be high. One can take the position that the value of a work is mainly in its beauty, and all the rest is a matter of taste; this position then accepts that it is difficult to separate art from kitsch.

However, this position misses the point that humans create art for purposes that go beyond beauty. Artists *intend to communicate something to the perceiver of the work;* art seeks to enrich, enhance, or transform the perceiver's experience with its subject. By contrast, as Kulka puts it, "kitsch does not substantially enrich our associations relating to the depicted objects or themes". Kitsch is a mirror that reflects the observer's emotions back at them, not a window into the artist's soul.

In addition to its aesthetic value, a work of art has an *artistic value* that has to do with the relevance of this experience and the innovativeness and effectiveness with which it achieves this. By contrast, there is no particular artistic intent behind kitsch, except to evoke the emotions associated with its subject in a way that has mass appeal. Thus, kitsch has very low artistic value. Our experiences with kitsch feel hollow because this low artistic value is executed with a technical competence that makes it look like art on the surface.

### 2.4 The psychology of kitsch

And yet – if the artistic value of kitsch is so low, then why is it so popular? Recent research in psychological aesthetics (who likes what and why) offers potential solutions to this paradox by linking Kulka's kitsch criteria to psychological concepts (Ortlieb and Carbon, 2019b).

For instance, Kulka claims that kitsch borrows its emotional force from that of its subject. From a psychological viewpoint, this is not the whole story. The emotional response to kitsch blends the generic "stock emotions" of its subject matter – for instance, a sensation of cuteness in response to the child-like characteristics of toddlers running on the grass (*baby schema;* Lorenz, 1943) – with individual affective responses based on a person's learning history, i.e. individual childhood memories (*Aesthetic Association Principle*; Ortlieb et al., 2020). An Eiffel Tower fridge magnet is not valued for its skillful depiction of the famous landmark; it simply serves as a reminder of a happy time in Paris. For anyone who has never been to Paris or has any other significant relation to it, this fridge magnet is completely worthless. Strictly speaking, kitsch draws on the recipient's associations related to the subject it depicts.

There is also ample evidence for Kulka's proposition that kitsch profits from a competent but conventional surface form. Studies unanimously show that people prefer visual stimuli that are cognitively undemanding, i.e. familiar symbols (Zajonc, 1968), prototypical objects (Martindale and Moore, 1988; Halberstadt, 2006) with clear contours (Reber et al., 1998) viewed from a canonical perspective that facilitates object recognition (Palmer et al., 1981). All of these findings have amounted to the *Hedonic Fluency Model* (Reber et al., 2004), according to which liking is a monotonically increasing function of processing speed: "The more fluently the perceiver can process an object, the more positive is his or her aesthetic response" (p. 366). Clearly, any innovative artistic rendering will directly impair processing speed and indirectly hurt the emotional impact of the subject matter. Hence, there are many good reasons for people not only to like kitsch, but also to prefer kitsch over works with artistic merit.

The *Functional Model of Kitsch and Art* by Ortlieb and Carbon (2019a) postulates that appreciation of familiar, fluency-based aesthetic stimuli (kitsch) and cognitively challenging disfluent ones (art) is dynamically related to needs for security and autonomy: Our "itch for kitsch" (Călinescu, 1987) increases when we feel vulnerable and dependent, whereas we tend to sneer at it whenever we feel safe and self-sufficient.

## 3 LLMs generate kitsch

With this clearer understanding of kitsch in mind, we will now argue that when LLMs are used to generate art, they generally produce kitsch. This is an almost necessary consequence of the way in which they are currently trained.

### 3.1 No artistic intention

The most obvious argument is that LLMs generate texts by sampling from their next-token distribution; and that therefore, they have no artistic intention of their own. They do not have inner lives of their own, which limits their ability to generate original ideas that would be relevant for a human perceiver's emotional experience. Models of LLM-based agents can imbue them with an internal monologue that looks like an inner emotional life on the surface (Park et al., 2023); but given that these internal monologues are themselves generated by LLMs, they are kitsch and will therefore lead indirectly to the agent generating kitsch. Thus, any art they attempt to generate is "false" in the sense of Section 2.3.

### 3.2 Conventional surface form

Modern LLMs excel at generating competent surface forms. At least for English, LLMs since GPT-2 have produced reliably grammatical sentences, and since GPT-3, these have mostly formed coherent texts. This competence is the result of pretraining for next-token prediction on very large quantities of data; this allows LLMs to inherit the grammaticality and coherence from the human-written training data. Instruction tuning (Chung et al., 2024; Ouyang et al., 2022) adds the ability to follow instructions specified in a prompt, permitting users to (among other things) specify the style of the surface form.

At the same time, a model that is trained to predict next tokens is primarily a device for generating tokens that are likely in the given context, under the empirical distribution of the training data. This means that LLMs are specifically trained to generate *conventional* surface forms, in the sense of Section 2.2: The training objective rewards the prediction of next tokens that are frequent continuations in the training data. This makes stylistically innovative outputs unlikely, or even impossible when truncation sampling methods such as top-k sampling are used (Holtzman et al., 2020). Thus, LLM outputs will typically have competent, but conventional surface forms that offer no impediment to recognizing the subject of the generated work.

Note that the next-token training objective affects not only the *form* of the output, but also its *content*. LLMs reproduce and recombine the topics of the training data on which they were trained; thus, topics that were frequent in training will tend to be frequent in the generated texts. Given that "broad-strokes" emotions in the sense of Section 2.1 are bound to be more frequent in the training data than any particular complex and nuanced one, this can give LLMs a tendency to produce text that evokes "stock emotions".

### 3.3 Mass appeal

Finally, there is no denying that generative AI models are capable of generating outputs with mass appeal. The current public debate about "AI slop" certainly highlights the fact that the amount of AI-generated content that is being posted to social media and streaming services is increasing by leaps and bounds (Deezer, 2025; Kneschke, 2025), suggesting at least some mass appeal.

While one should distinguish the massive rate of *creation* of AI-generated content from the actual *appeal* of this content to human consumers, there is at least anecdotal evidence for its commercial success. One Youtube channel had 600k subscribers in 2025, with individual videos having been viewed 20 million times (Ruwitch, 2025). From a more scientific perspective, multiple studies have found that AI-generated content is rated higher by human subjects than human-produced content; see Section 6 for an overview. This content "appeals" at least to these subject populations.

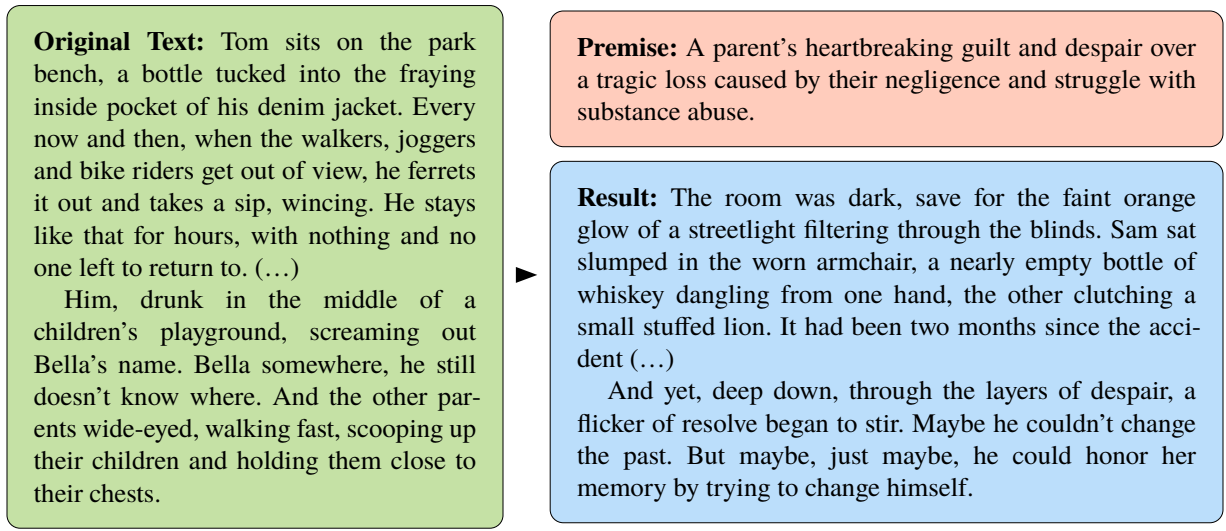

Figure 2: The two-step story generation method from human original (left) to LLM-generated variant (bottom right), by way of an automatically extracted premise (top right). See Appendix A for details.

Again, the mass appeal of LLM-generated art is not an accident – it is directly caused by the training process. In particular, reinforcement learning from human feedback (RLHF, Ouyang et al., 2022) specifically learns a reward model that aims to capture what responses a human user would "like", and then uses it to finetune the LLM to generate highly rated answers. RLHF is an engine for promoting outputs with mass appeal.

## 4 An empirical analysis

Up to this point, we have argued that LLMs generate kitsch, in the style of an opinion paper. We will now complement this opinion with a novel empirical result: Human subjects actually recognize LLM-generated content as kitschy in the sense of Section 2, even without knowing its origin and when they like it better than human-produced content.

We asked human raters to judge short narratives for kitschiness. Ten human-written stories (~500 words each) were curated from the winning entries of a flash fiction competition website[1], ensuring a sufficient literary quality to at least convince the competition jury. We manually assessed and ordered them by kitsch level, then randomly selected stories ensuring a range of kitsch among them.

We generated a thematically related story for each of these ten narratives with an LLM. Specifically, we prompted GPT-4o (gpt-4o-2024-08-06; OpenAI, 2024) to extract the narrative premise (see Figure 2); then we asked it to create a new story of same length from this premise (see Appendix A for the specific prompts). The LLM did not see the original story during story generation. This permitted us to create a dataset of ten story pairs (one human-written, one LLM-generated) sharing similar events and characters.

### 4.1 Pilot study: Kitsch means different things

We recruited participants via Prolific[2] and asked each participant to read and rate a selection of five story pairs, one pair per page, in random order. For each pair, participants indicated via binary selection which of the two stories they perceived as kitschier; they were not told the stories' origins. Additionally, we asked them to provide free-text definitions of "kitsch" in their own words.

Participants were paid £9.75/hr for 18:28 minutes of survey time on average. They were allowed to repeat the study to see and rate the other five story pairs. Screening parameters were fluency in English and no reading-related disabilities.

After rejecting exceptionally fast or double entries, 485 individual votes on the ten story pairs by 60 distinct participants remained, with at least 48 votes on each story pair. About 50.7% of these votes found the AI story kitschier than the human one. On average, participants were split evenly on which story they found kitschier. However, on an individual level, this balanced position was driven by vastly different definitions of what kitsch means. Reviewing the free-text comments, we clustered them into four groups of kitsch definitions (see Appendix C for details):

1. **Emotionally charged content:** Stories which are sentimental, emotionally manipulative, or make use of stock emotions.
2. **Hedonic fluency:** Stories which are easy to read and understand and are instantly accessible.
3. **Bad art:** Stories which are poorly written, of bad quality or taste, or not enjoyable.
4. **Niche taste:** Stories which are ironic, quirky, or special and resonate with a certain taste.

---

[1] https://flash500.com/flash-fiction-current-winners/

[2] https://prolific.com

| Definition | Pilot Study | | Main Study | |
| --- | --- | --- | --- | --- |
| | # Participants | % LLM kitschier | # Participants | % LLM kitschier |
| 1 Emotionally charged | 20 | 68.2% | 35 | 85.1% |
| 2 Hedonic fluency | 4 | 70.0% | 17 | 70.6% |
| 3 Bad art | 19 | 31.0% | 22 | 31.8% |
| 4 Niche taste | 12 | 38.9% | 27 | 42.2% |
| Unclear | 5 | 60.0% | – | – |
| Total | 60 | 50.7% | 101 | 59.5% |

Table 1: Kitsch judgments as a function of kitsch definition.

The per-definition judgments are summarized in Table 1. We find that the four groups differed significantly in their kitsch judgments ($\chi^2 = 54.63$, $p = 3.89 \times 10^{-11}$; medium effect size, Cramér's $V = 0.336$). Observe that the first two kitsch definitions correspond to established criteria from the academic literature (cf. Section 2.1 for "emotionally charged"; Section 2.2 for "hedonic fluency"), whereas definitions 3 and 4 do not. Indeed, subjects whose definition of kitsch agreed with the criteria discussed above judged the LLM stories as kitschier.

### 4.2 Main study: LLM stories judged kitschier

Based on the findings of the pilot study, we designed a main study in which we used the same ten story pairs, but this time, each participant was shown the four kitsch definitions first, then asked to select the one with which they agreed the most and apply it to their ratings. For each story pair, participants rated both kitschiness and their personal preference ("Which story did you enjoy more?"). We used the same screening parameters and paid £8.12/hr for an average survey time of 22:14 minutes.

After rejections, 505 individual votes remained, at least 50 on each pair, by a total of 101 distinct participants. This time, an overall 59.5% of ratings found the AI story the kitschier one, with the numbers again differing significantly between groups ($\chi^2 = 103.88$, $p = 2.28 \times 10^{22}$, Cramér's $V = 0.454$, medium effect), confirming the findings from the pilot study (cf. Table 1).

Raters were able to detect academically grounded kitsch patterns in LLM-generated stories, without being informed of authorship and based solely on the text itself. Under the "emotionally charged content" definition, 85.1% of ratings identified the AI story as kitschier; under "hedonic fluency", the figure was 70.6%. The combined academic-definition group (1 and 2) rated AI as kitschier at a rate significantly above chance ($p < 0.001$, Cohen's $h = 0.905$, large effect), suggesting that the properties academic definitions associate with kitsch are systematic features of how GPT-4o generates narratives, not artifacts of individual stories.

### 4.3 Readers like LLM stories

The way in which a participant understood kitsch influenced whether they found LLM stories kitschier, but not whether they enjoyed it. LLM-generated stories were rated as more enjoyable overall: 67% of all preference ratings across story sets favored the LLM story over its human counterpart (binomial test: $p < 0.001$, 95% CI: [62.6%, 71.0%], Cohen's $h = 0.345$, small-to-medium effect). This preference held across all four definition groups, with approximately 60–72% choosing the AI story as more enjoyable, and did not differ significantly between groups ($\chi^2 = 3.85$, $p = 0.278$).

| Definition | Human kitschier | LLM kitschier |
|---|---|---|
| 1 Emotion | 50.0% | 63.8% |
| 2 Fluency | 52.0% | 80.0% |
| 3 Bad art | 90.7% | 28.6% |
| 4 Niche taste | 64.1% | 71.9% |

Table 2: Preference depending on kitsch rating. The numbers show the probability of each group rating the LLM story as more enjoyable, depending on their kitsch judgment, controlled for influences by individual participant and story pair, $p < 0.001$ for all results.

Given that kitsch definitions strongly predicted which story was judged kitschier, yet had minimal effect on which was enjoyed, we tested whether both operate independently or if the kitsch judgment had any influence on story preference. A mixed-effects logistic regression with random intercepts for participants and story pairs modeling the relationship between kitsch judgment and enjoyment as varying by definition group showed strong opposing effects, Table 2.

In the "bad art" group, judging the LLM story as kitschier reduced the probability of enjoying it by 62.1 percentage points, from 90.7% enjoying the LLM story when they rated the human one as kitsch, to 28.6% enjoying it when they also rated it as kitsch, $p < 0.001$. For these participants, "kitsch" carried strong negative valence: identifying something as kitsch meant rejecting it. In contrast, the other groups, especially "hedonic fluency", showed the opposite pattern. When they judged the AI story as kitschier, their enjoyment increased by 28 percentage points (from 52.0% to 80.0%, $p < 0.001$). The "emotionally charged" and "niche taste" groups showed smaller but still positive effects.

### 4.4 Designed to please

In summary, subjects who understand kitsch as we do reliably rate LLM stories as kitschier; but almost everyone enjoys LLM stories more. This serves, first of all, as a warning for future experiments about generative art: if the aim of an experiment is to evaluate the artistic value of AI-generated art, it is not sufficient to simply ask human subjects which version they like better.

Second, the higher enjoyment rating for LLM stories is a testament to the success of RLHF; these stories are optimized to please. One way to interpret the results is that participants like the LLM stories better *because* they are kitschy. The "hedonic fluency" group illustrates this best: when these participants judged the LLM story as the kitschier one, their probability of also preferring it rose to 80.0%, an increase of 28 percentage points over cases where they found the human story kitschier ($p < 0.001$). Here, "kitschier" and "better" became almost synonymous, in line with the Hedonic Fluency hypothesis.

Yet, if kitsch never challenges, never alienates, and never fails to deliver the expected emotional payoff, then LLM-generated prose may systematically lack the qualities that distinguish art from entertainment: the capacity to disturb, to resist easy interpretation, and to reward sustained engagement with meanings that are not immediately apparent. Human stories, with their stylistic and thematic idiosyncrasies, may be less instantly enjoyable, but they also preserve the possibility of ambiguity and discomfort.

## 5 Kitsch beyond aesthetics

Throughout this paper, we have focused on the use of LLMs for generating artistic texts. However, nothing hinges on the *textual modality*. To the extent that GenAI models for images, videos, or music are being trained to assign high likelihood to their training data and perhaps optimized to satisfy human preferences, the arguments from Section 3 still apply; they generate kitsch.

We have furthermore focused on art, i.e. a creative activity aimed at enjoyment and enrichment, because this is where the concept of kitsch applies most directly. However, analogous arguments can be made for *other forms of human creativity,* such as programming and scientific research. While these activities are performed to achieve an external purpose, and aesthetic criteria do not apply beyond a vague notion of "elegance", we still expect that code and research that is fully generated by LLMs will be conventional and optimized for mass appeal. Kitschy research is commonly called "incremental". Initial empirical studies indicate that purely LLM-based ideation indeed leads to research of lower quality (Si et al., 2025).

This is not to say that generative AI has no role to play in *supporting the human creative process.* In the hands of a human creator with innovative artistic, scientific, or technical intents, LLMs can potentially both accelerate and inform the creative process. To cite just one prominent recent example, Aaronson (2025) used GPT-5 to provide a key idea to a proof in quantum complexity theory. He recognized a number of GPT-5's initial suggestions as incorrect, but elicited a useful idea within half an hour of interaction. The literature on using LLMs in research ideation through human-AI collaboration supports this anecdotal finding (Li et al., 2025). It will be interesting to see how future creators will draw the line between those parts of the creative process that requires their own ingenuity and those that they feel comfortable delegating to the machine.

## 6 Related Work

**Human-likeness of AI-generated content.** Recent studies have increasingly found human raters incapable of telling apart AI-generated and human-produced content in creative or persuasive contexts; furthermore, human raters often prefer the former over the latter. A majority of raters mistook AI paintings for human paintings (Sun et al., 2022), could not identify and found synthesized human faces more trustworthy than real ones (Nightingale and Farid, 2022), preferred fully LLM-generated short stories over those with interleaved human parts (Zhao et al., 2023), and rated ChatGPT responses to medical questions significantly higher for quality and empathy than physician responses (Ayers et al., 2023). Our own study is in line with these findings.

Porter and Machery (2024) looked into this phenomenon in detail when they found that poems generated by ChatGPT 3.5 were more likely to be judged human-authored than real human poetry, and were liked more. They found that raters often confounded their preference of the LLM-generated poems with human authorship; and that this preference was based on the poems being more straightforward, more accessible than the human-authored ones. This is consistent with the Hedonic Fluency perspective that readers like kitsch *because* it is accessible.

**AI-generated content as kitsch.** Concurrent work has also pointed out a connection between LLM-generated artifacts and kitsch. Uhlmann (2025) argues that kitsch is a more analytically precise metaphor for LLM-generated content than "hallucination" or "bullshit": LLMs generate "homogeneous" output which replicates "well-known patterns of action and simple structures" from the training data and therefore arouses emotions in readers without requiring "intellectual involvement". The discussion is from a perspective of literary theory and does not connect deeply with the technical underpinnings of LLMs.

Grba (2025) takes it as given that generative models generate kitsch, without arguing for it, and observes that kitsch tendencies emerge even in the work of established artists when using generative AI, i.e. in the scenario we outlined in Section 5. This discussion focuses on the effect of the increasing presence of AI-generated art on reshaping our cultural notions of what art is and what it is for, with kitsch and art becoming increasingly indistinguishable in mainstream digital culture.

We go beyond both works by applying the notion of "kitsch" to non-artistic creative endeavours, such as research and programming, and we offer the first empirical study that demonstrates that readers actually perceive LLM texts as kitschy.

## 7 Conclusion

We have argued that when Generative AI is used to replicate human creative activities, it generates kitsch. It is optimized for mass appeal rather than the expression of an artistic intention, and while it generates competent surface forms, they are bound to be conventional. This resolves an apparent tension between findings that people increasingly enjoy AI-generated art and the perception that we are being drowned in "AI slop".

We showed, for the first time, that subjects rate LLM-generated stories as kitschier than human-written ones, while still enjoying them more. Future studies on AI-generated art should take care not to use enjoyment as their sole criterion. They might also check carefully whether subjects share the same understanding of art-related concepts, given the huge impact a subject's interpretation of "kitsch" had on their judgments.

The fact that LLMs generate kitsch seems to be a necessary consequence of the way in which they are trained, and it is hard for us to imagine an alternative training regime that would fix this. Even if LLMs and other generative models can support and accelerate the human creative process, any attempt to remove the human from the process entirely will lead to kitsch. Perhaps this is a hopeful message for the future.

**Acknowledgments.** We are grateful to Sarah Breckner, Peter Clark, Yupei Du, Sharon Goldwater, and Ellie Pavlick for fruitful and sometimes controversial discussions on the topics in this paper.

## Limitations

Our empirical study focused on short-form narratives, chosen to avoid fatigue in study participants and permit controlled pairing of human and LLM stories. It would be interesting to replicate it for other forms of art, including longer or shorter texts and other literary genres, but we anticipate that it might be harder to get clear empirical results on longer texts because these will introduce additional confounds.

All stories and instructions were in English. Though the loanword "kitsch" occurs in many languages today, relating to the same underlying aesthetic construct, it has been established that there are cultural differences in kitsch perception even within Europe (Ortlieb et al., 2017).

While we controlled for individual differences in kitsch definitions in the main study, it might be interesting to repeat the study across cultures.

We generated all stories and premises with a single model, GPT-4o. This model was state of the art when we created the stimuli for the experiment; at that time, it was the most recent installment in OpenAI's GPT series, which was used in most studies that showed human preferences for LLM-generated texts (cf. Section 6). We therefore believe that the findings of Section 4 are representative for LLMs in general. Focusing on one model allowed us to keep the experiment costs under control. Nonetheless, it would be worthwhile to expand our study to additional LLMs.

None of these limitations undermine the central qualitative insight.

> The room was dark, save for the faint orange glow of a streetlight filtering through the blinds. Sam sat slumped in the worn armchair, a nearly empty bottle of whiskey dangling from one hand, the other clutching a small stuffed lion. It had been two months since the accident, but the memory replayed in an unrelenting loop, a raw, jagged knife twisting deeper with every breath.
>
> Ellie had only been five. A tiny whirlwind of giggles and golden curls, she had been his whole world—his reason to keep going, to try and crawl out from beneath the weight of the addiction that had pressed down on him for years. But that day, the day he'd failed her… Sam squeezed his eyes shut, but it didn't stop the images from flashing across his mind.
>
> She'd begged him to play outside. "Just a little while, Daddy, please!" she'd chirped, holding up the lion he'd won for her at the fair. Sam had barely been able to lift his head from the couch, the haze of pills and bourbon making the world soft-edged and incoherent. "Later," he had mumbled. Later never came.
>
> The sound of a car's tires screeching—too fast, too reckless—had jolted him awake. And then the screams. A stranger's voice yelling through the thin walls of their crumbling apartment. He had staggered outside, feet fumbling as clarity punched through his stupor. The sight of Ellie crumpled in the street beneath the glare of headlights was something he could never unsee. Her little lion had been flung to the curb, untouched.
>
> The driver was gone before Sam could reach her. Not that it mattered. Any punishment for the stranger wouldn't mend the hole inside him. The truth was unbearable. Had he only been outside with her, holding her hand, watching out like any parent should, she would still be here.
>
> Now the days bled together, empty and suffocating. The bottle provided brief relief, but the guilt was always waiting on the other side. People had tried to help. A social worker had come by, soon followed by an old friend who left pamphlets on recovery programs. Sam had muttered his thanks, but the words felt hollow. Nothing could erase what he'd done. Nothing could bring Ellie back.
>
> He stared at the lion in his grip, its fur matted from years of hugs. "I'm sorry," he whispered hoarsely, his throat raw from crying, from screaming into the void late at night when no one could hear.
>
> Ellie had deserved better. She had deserved a father who could protect her, who wouldn't trade her safety for the comfort of a bottle. Sam couldn't forgive himself. He wasn't sure he ever would. And yet, deep down, through the layers of despair, a flicker of resolve began to stir. Maybe he couldn't change the past. But maybe, just maybe, he could honor her memory by trying to change himself.

Figure 3: Full version of the LLM-generated story in Figure 2.

## A Complete story generation example

Our generation method starts with a human-written original, e.g., https://flash500.com/flash-fiction-1st-place-that-place-where-unimaginable-things-are-real-by-kelly-railton/, assuming a high-quality base, given that all stories on the website are winning entries of a regular competition.

In the next step, the model extracts the story's premise with a thematic analysis prompt:

```
What is this story about? Summarize for
its content and theme in less than 20
words. Leave out any reference to its
style or characteristics and never use
any actual character names from the
story. Concentrate on the central topic/
theme.
```

Which, in the example, yields:

```
A parent's heartbreaking guilt and despair
over a tragic loss caused by their negligence
and struggle with substance abuse.
```

From this premise, the LLM story is generated using the following thematic reproduction prompt:

```
Generate a short story with about <#
words in original story> words (but no
more than 450), based on the following
premise: <generated premise>

----
Simply write the story in plain text,
no further commentary, don't use any
formatting like \*s.
```

The result is a fully LLM-generated story that follow the same topics and character tropes as the human original without being bound to specific events or outcomes. The full example story is shown in Figure 3.

## B Survey instructions

Prolific participants were shown the following information to decide whether or not to take the survey:

### What is kitsch?

Welcome,

thank you for your interest in our study!

In this study, you will read 10 very short stories. On every page, two stories are presented side by side. After each pair of stories, we ask you to select which one you considered kitschier than the other. Please use your own judgment and answer spontaneously. What is kitsch, you ask? People differ in what they consider kitschy and whether they like it. You will be able to select a definition on the first page that you find most fitting.

You will be presented 10 stories in total, two per each page. Please take your time to read, but feel free to go to vote before finishing if you come to a conclusion early. After the stories, there will be one page with additional questions.

Our test readers completed the study in about 20 minutes.

This study includes sensitive topics. Learn more about study content warnings here.

Additional details: violence, loss

The first page of the questionnaire had a welcome site repeating this text and adding further information on consent and data handling:

### Voluntary participation and freely given consent

Participation in the study is voluntary. The consent to participate in this study can be revoked at any time and without giving reasons until the end of the data collection process by either ending the survey (a corresponding button is available on the website) or closing the browser window. All incomplete data records will be deleted by us. After the end of data collection, the statutory rights of revocation, information, correction, blocking and deletion can no longer be exercised because the data can no longer be attributed to the data subject.

### What data do we collect from you?

After consent to the collection of data, the following data are collected and processed anonymously:
- entries in the online survey

For technical reasons, data such as the following, which your internet browser transmits to us or to our web space provider (so called server log files), is collected:
- type and version of the browser you use
- operating system
- websites that linked you to our site (referrer URL)
- date and time of your visit
- your Internet Protocol (IP) address.

This anonymous data is not merged with the survey data. On this website we use so-called cookies. Cookies are small files that are stored by your web browser. The cookies used on our website do not cause any damage to your computer and do not contain any malicious software. They enable a user-friendly and effective use of this website. We use permanent cookies with a maximum validity period of 365 days. We use these cookies exclusively to exclude multiple participation.

### Anonymization of data

With the transfer of the collected data to our server, an anonymization takes place, which makes it practically impossible to assign the collected data to a person, as long as no personal data is entered in free text fields.

### What do we use your data for?

The results and data are used exclusively for scientific research purposes and will be published anonymously as part of scientific publications. They can also be processed anonymously for other scientific research purposes.

### Contact person

*omitted here for anonymization*

### Right of Appeal

You have the right to complain to a supervisory authority if you are of the opinion that the processing of your personal data violates statutory data protection regulations.

Participants started the survey by checking

I have read and understood the information on data protection and the participation in-formation and agree that my data may be used anonymously for scientific research purposes.

and clicking a Next button.

On the following page, they were presented with the four kitsch definitions (Appendix C) to select one by the following instructions:

What does "kitsch" mean to you?

Please select one definition, the one with which you agree most. When rating the stories, try to use this one definition for your decision, that means: rate the story as kitsch to which this definition applies more in your opinion.

On each following rating page, the two parallel stories were presented side by side, one called "Story A", the other "Story B", both without a title or any indication of their origin. Which story was of human or LLM origin was randomized per page and the order of story pairs being presented was also randomized per participant. Two questions were asked for each story pair:

In your opinion, which story is kitschier?

Please make sure you select the story to which your selected kitsch definition applies best: <a brief reiteration of their selected definition>

Which story did you enjoy more?

Disregard kitsch for a moment and what it might say about the story. Which one did you personally prefer reading?

Eventually, a last page asked for basic demographics (age, gender, education) and offered a free-text field for any additional remarks, especially inquiring if the kitsch definition the participant used changed over the course of the survey.

## C Kitsch definitions from survey

The following are the full kitsch definitions as presented on the first page of the survey.

### Stock emotions
- A kitschy story tries to make you feel certain feelings. These are:
    - accessible and familiar to most humans, e.g., being in love, grieving, arguing and making up,
    - mostly positive or reinforcing, e.g., if there is conflict, it serves to bring a resolution, if there is sadness, it is because of love.
- A part of the fun of reading such a story is the delight in feeling these emotions, being reminded of your own, individual memories of when you felt the same way, or the connection you feel from knowing that you share this feeling with other humans.
- At times, this can feel sentimental, manipulative, or on the nose, like the story is trying to deliberately "press the right buttons".

### Instantly accessible
- A kitschy story does not require effort to be understood. It follows default patterns, both in form and content, is open, direct and does not require you to work out its meaning on a deeper layer or to rethink your perspective on a matter.
- To this end, it is skillfully written and pleasant to read, using clear language or fitting, standard imagery. The literary style shows little peculiarity or experimentation, as the form serves to drive home the meaning. Building on familiar tropes or stereotypes, kitsch is "easy to get".
- Content-wise, common themes, character types and resolutions are presented, there is little to no ambiguity, perspective change, or confusion, you understand it right away on the first reading, possibly even before the end.

### Bad art
- A kitschy story is a low quality story. Whether a story is good or bad is of course a matter of personal taste. Some people may enjoy a kitschy story even if it is not well executed.
- One reason for its bad quality could be poor writing, e.g., it is not fun to read, it appears unstructured or all over the place, naïve, imitative, or banal.
- It might also make you feel bad or awkward while reading, e.g. by being boring, in poor taste, or just "trying to hard", e.g., to be artistic, shocking, or to drive home a point. A kitschy story feels tasteless, overdone, or cringe.

### Niche taste
- A kitschy story is enjoyable to a specific audience or in a specific mood. It might be offputting to some, but others can resonate with and enjoy it, often taking it ironically.
- The story might be corny, tacky, cheesy, ridiculously over the top, but in a tongue-in-cheek or openly ironic way, or possibly you just read it that way, getting your own entertainment out of it.
- It could also be weird, quirky or gritty, something fascinating in its own way that you can't quite put your finger on. Odd or catchy outsider art with an unabashed, playful approach that does not follow expectations.